\documentclass[runningheads]{llncs}
\usepackage[T1]{fontenc}
\usepackage{graphicx}
\usepackage{amsfonts}
\usepackage{amsmath}
\usepackage{makecell} 
\usepackage{multirow} 
\usepackage{graphicx} 
\usepackage{booktabs} 
\usepackage{siunitx}  
\usepackage[table]{xcolor} 
\usepackage{hyperref}

\usepackage{tikz}
\usetikzlibrary{shapes.geometric, arrows.meta, positioning, shadows}
\usepackage{bm}

\begin{document}

\setlength{\textfloatsep}{10pt plus 2pt minus 2pt}  
\setlength{\floatsep}{8pt plus 2pt minus 2pt}       
\setlength{\intextsep}{10pt plus 2pt minus 2pt}     

\title{Semantic Pareto-DQN: A Multi-Objective Reinforcement Learning Framework for Financial Anomaly Detection}
\titlerunning{Semantic Pareto-DQN: A Multi-Obj. RL for Financial Anomaly Detection}
%
\author{Cláudio Lúcio do Val Lopes\inst{1}\orcidID{0000-0003-1655-2283} 
\and \\ 
Lucca Machado da Silva\inst{1}\orcidID{0009-0002-7191-6756}
}

\authorrunning{Cláudio L V Lopes and Lucca M. da Silva}
%
\institute{A3Data  - \url{http://www.a3data.com.br}\\
Correspondence to: \email{claudio.lucio@a3data.com.br}\\
 }
\maketitle              
\begin{abstract}
Financial anomaly detection suffers from extreme class imbalance, causing traditional single-objective algorithms to exhibit ``fraud collapse'', defaulting to the majority class and failing to balance anomaly interdiction with customer friction. To overcome this without distortive data resampling, we propose the Semantic Pareto-DQN, a multi-objective reinforcement learning framework. Our approach synthesizes heterogeneous transaction features into cohesive natural-language narratives, encoded by large language models, thereby producing a robust, scale-invariant state representation. The agent optimizes a vectorial reward that explicitly decouples financial efficacy, operational friction, and semantic discovery. By mapping the continuous Pareto frontier, the system dynamically navigates the asymmetric costs of missed anomalies versus false positives. Empirical evaluations across E-Commerce fraud and UCI Credit datasets show that semantic Pareto-DQN successfully shatters the zero-recall trap. It achieves superior minority-class recall compared to scalarized baselines, providing an alternative to trade bounded operational friction for financial anomaly discovery.

\end{abstract}
\section{Introduction}

Financial anomaly detection, encompassing e-commerce fraud and credit card defaults, is characterized by extreme class imbalance and adversarial adaptation \cite{jtaer20020121,11172851}. Some machine learning approaches, including deep artificial neural networks, has been deployed to maximize predictive accuracy and safeguard data privacy in fraud detection \cite{sulaiman2022review}. However, as highlighted in recent reviews, these standard supervised algorithms predominantly optimize a single scalar metric, typically maximizing the log-likelihood. 
In e-commerce environments characterized by extreme class imbalance, this singular mathematical focus structurally incentivizes the model to collapse onto the majority-class manifold. Consequently, these static classifiers are fundamentally ill-equipped to dynamically balance the non-linear trade-offs between maximizing anomaly detection efficacy and minimizing customer friction.


To mitigate this extreme class imbalance, traditional literature frequently relies on data-level resampling techniques (e.g., SMOTE or ADASYN) \cite{hassanat2022stop}. However, as detailed in Section \ref{Background}, these methods often introduce fictitious samples that distort real-world decision boundaries. Avoiding these distribution alterations, our methodology focuses purely on an algorithmic intervention.

Overcoming the static nature of standard supervised classifiers, recent literature has formulated fraud detection as a sequential decision-making process governed by Reinforcement Learning (RL) \cite{Tang2023,cao2025driftshield,papanastassiou2025reinforcement}. By treating transactions as state transitions within a Markov Decision Process (MDP), RL agents learn a dynamic policy $\pi_\theta$ that maximizes cumulative expected returns rather than merely minimizing instantaneous classification error. However, the operational goal of an enterprise fraud detection system is fundamentally multi-dimensional: rather than minimizing all statistical errors, it requires dynamically managing the asymmetric trade-offs between maximizing the interdiction of rare fraud (True Positives) and bounding the inevitable operational friction (False Positives) associated with high-sensitivity detection. 
Current state-of-the-art architectures attempt to navigate these constraints via scalarization and asymmetric reward shaping \cite{cao2025driftshield}. In adversarial financial environments, this \textit{a priori} scalarization restricts the agent to the convex hull of the Pareto front, preventing it from discovering the true optimal trade-off surface. This fundamental limitation motivates a critical transition toward a Multi-Objective Reinforcement Learning (MORL) framework that can navigate competing operational goals natively.

Traditional single-objective supervised models, as well as standard Deep Q-Networks (DQN), typically optimize purely for aggregated accuracy or immediate financial return \cite{papanastassiou2025reinforcement}. In highly imbalanced environments (often $<10\%$ fraud), this structural bias invariably induces ``fraud collapse'', a phenomenon where the architecture overfits to narrow, high-density historical heuristics. Consequently, single-objective models become ineffective at dynamically navigating the antagonistic, non-linear trade-offs between minimizing financial loss and discovering novel attack vectors.

To circumvent the limitations of static scalarization, we formalize sequential financial risk detection as a Semantic Multi-Objective Markov Decision Process (MOMDP) \cite{Rehman2025}. We introduce a MORL framework utilizing a Pareto Deep Q-Network (Pareto-DQN) \cite{Hayes2022}. Rather than optimizing a single scalarized target, our agent explicitly decouples the objectives into a vectorial reward signal $\bm{r_{t}}=[r_{eff}, r_{drift}, r_{div}]^{\top} \in \mathbb{R}^{3}$, ensuring that financial efficacy, customer friction, and semantic anomaly discovery are treated as distinct mathematical entities. By employing hypervolume-based action selection to navigate the continuous Pareto frontier \cite{reymond2019paretodqn}, the agent balances financial efficacy with proactive discovery of semantically distant fraud topologies, thereby effectively disrupting the positive feedback loops that drive algorithmic collapse.

To test our framework's capacity to maintain a detection manifold, we conduct empirical evaluations across two distinct financial databases. First, we utilize the E-Commerce fraud dataset \cite{grover2023fraud} to assess the Pareto-DQN's against extreme class imbalance and non-stationary behavioral drift. Second, we benchmark the architecture against the UCI default of credit card clients dataset \cite{default_of_credit_card_clients_350} to evaluate its generalizability in environments with moderate imbalance and differing temporal correlations. Empirical evaluations demonstrate that our MORL approach maps the Pareto frontier across both domains, achieving superior minority-class recall while sustaining high state trajectory variance to mitigate global recall collapse.

\section{Background}  \label{Background}

In the context of financial anomaly detection, recent literature has explored deep learning architectures to combat adversarial adaptation \cite{cao2025driftshield,8374720,10417044}. For instance, Sulaiman et al. \cite{sulaiman2022review} provide a comprehensive review of machine learning approaches for credit card fraud detection, emphasizing the integration of neural networks within federated learning frameworks to improve classification accuracy while preserving data confidentiality. While their work underscores the efficacy of advanced neural architectures in identifying fraudulent patterns, these methods remain strictly anchored to single-objective optimization paradigms. In such traditional frameworks, auxiliary operational goals, such as minimizing user friction or the diversity of discovered fraud typologies, are either ignored or statically scalarized into a fixed loss function, e.g., $\mathcal{L} = w_1 \mathcal{L}_{fraud} + w_2 \mathcal{L}_{friction}$. 
\subsection{Multi-Objective Reinforcement Learning}


RL optimizes long-term operational goals in sequential tasks by modeling them as a Markov decision process driven by environmental feedback \cite{Afsar2022}. Within value-based deep RL pipelines, Deep Q-Networks (DQNs) are a prominent standard for discrete action spaces due to their robust update mechanisms. Empirical evidence shows that these interactive architectures outperform traditional static classifiers by autonomously adapting to dynamic, evolving behaviors \cite{Wang2022DRL}.

Despite these advances, traditional single-objective optimization presents a fundamental mathematical limitation. Real-world financial environments inherently involve multiple conflicting objectives requiring simultaneous optimization \cite{Rehman2025}. Standard deep RL typically struggles to balance predictive accuracy with non-accuracy metrics (such as customer friction), making it ineffective for navigating complex operational trade-offs \cite{Rehman2025}. Multi-objective reinforcement learning addresses this gap by using vector-valued rather than scalar rewards, enabling principled handling of conflicting objectives \cite{Hayes2022}. This approach is grounded in Pareto optimization, a theoretical framework in which improving one objective typically cannot be achieved without changing another \cite{Hayes2022}.

MORL represents the core mathematical engine of our proposal. The framework integrates high-fidelity semantic embeddings of transaction sequences into the anomaly detection pipeline. By employing a Pareto-DQN approach, the system addresses the limitations of traditional static classifiers and the narrow mathematical focus of single-objective optimization. This methodology establishes a paradigm for risk engines using the semantic latent spaces to navigate complex value trade-offs and map the optimal operational frontier, without exposing the platform to unchecked financial liabilities or unwarranted customer friction.


\subsection{Class Imbalance and the Pitfalls of Data Resampling}
The class imbalance problem in financial anomaly detection has been addressed through data-level interventions, algorithmic-level adjustments, or hybrid approaches. Data-level techniques employ oversampling algorithms, such as the Synthetic Minority Over-sampling Technique, SMOTE \cite{smote2002}, ADASYN \cite{ADASYN2008}, and Borderline-SMOTE \cite{han2005borderline} to synthesize new minority instances based on local feature topologies \cite{gosain2017handling}. Despite their widespread adoption to stabilize standard classifiers, recent critical reviews have exposed severe vulnerabilities in oversampling paradigms \cite{hassanat2022stop}. As noted by Hassanat et al. \cite{hassanat2022stop}, oversampling methods often generate fictitious minority samples that exhibit majority-class characteristics when projected into high-dimensional, real-world spaces \cite{hassanat2022stop}. Consequently, classifiers trained on such distorted topological manifolds suffer from degraded real-world generalizability and yield unreliable predictive posteriors. 

While traditional single-objective algorithmic modifications—such as focal loss or class-weighted adjustments—attempt to stabilize predictive posteriors, they restrict optimization to a static scalar metric. To bypass the mathematical pitfalls of synthesized data or single objective approach, our Semantic Pareto-DQN framework addresses the imbalance purely through an algorithmic-level intervention via MORL approach. This eliminates the need for data manipulation, showing that a precisely engineered reward topology can be addressed computationally in a way comparable to traditional oversampling.


\section{Proposed approach} \label{Proposed approach}

We present a Semantic MORL framework for financial risk detection that dynamically balances financial efficacy (fraud/default interdiction), customer friction (controlled false-positive management), and semantic anomaly detection. Empirical evaluations will be presented in Section \ref{Experiments}, in which traditional risk engines and single-objective reinforcement learning models are structurally predisposed to ``fraud collapse'' under extreme class imbalance; by rigidly over-optimizing for false positive suppression, these static models achieve high accuracy at the catastrophic expense of minority-class recall, frequently decaying to near-zero interdiction. 

Our approach dismantles this limitation by combining a narrative-based semantic representation space with a Pareto Deep Q-Network.
This allows the agent to dynamically navigate the inherent mathematical tension between detecting sophisticated financial anomalies, empowering the system to allocate operational friction where it yields the highest marginal gains in fraud detection.

\subsection{Semantic Embedding Pipeline}
To construct a robust, scale-invariant state representation that generalizes across diverse financial domains, ranging from instantaneous e-commerce transactions to longitudinal credit default histories, we eschew traditional tabular preprocessing (e.g., standard one-hot encoding and min-max scaling). Instead, we employ a Large Language Model (LLM) narrative encoding strategy. 

Let $\bm{x_t}$ represent the raw heterogeneous data of a transaction at time $t$. The raw features (categorical demographics, continuous financial amounts, and temporal sequences) are synthesized into a cohesive, target-free textual document $d_t$. For example, in the context of longitudinal credit assessment, $d_t$ encapsulates the client's demographic profile along with a 6-month sequential behavioral trajectory of billing statements and payment statuses. We map this narrative to a latent vector space using the \texttt{all-MiniLM-L6-v2} Sentence Transformer \cite{Reimers2019SBERT}:

\begin{equation}
    \bm{v_t} = \frac{\text{Encoder}(d_t)}{\|\text{Encoder}(d_t)\|_2}
    \label{eq:semantic_embedding}
\end{equation}

This architecture maps the textual narrative to a dense, 384-dimensional vector optimized for semantic similarity. The $L_2$ normalization is enforced as a critical geometric constraint; it ensures that the inner product between any two historical embeddings natively yields their cosine similarity, which is strictly required for calculating the spatial diversity reward during agent training. 

This semantic transformation provides some advantages. First, it inherently captures complex sequential interactions and contextual nuances that are frequently lost in sparse, high-dimensional tabular arrays. Second, leveraging a pretrained semantic space enables zero-shot generalization to unseen categorical values, mitigating the cold-start problem common in financial data streams. 

\subsection{Multi-Objective MDP Formulation}
We formalize sequential financial risk detection as a semantic MOMDP, defined by the tuple $(S, \mathcal{A}, \mathcal{P}, \mathcal{R}, \gamma)$. We utilize a semantic encoder to map $\bm{x_t}$ into a latent vector $\bm{v_t} \in \mathbb{R}^d$. Details are defined as follows:

\begin{itemize}
    \item \textbf{State Space ($S$):} In real-world fraud detection, the agent's localized operational friction is partially observable. To resolve this POMDP, the state $\bm{s_t} \in S$ is constructed by concatenating the current transaction embedding $\bm{v_t}$ with a continuous, rolling measurement of the agent's recent False Positive Rate ($FPR_t$). Thus, $\bm{s_t} = \left[ \bm{v_t} \parallel FPR_t \right] \in \mathbb{R}^{d+1}$, ensuring the agent possesses awareness of both the semantic transaction and its current friction budget.
    
    \item \textbf{Action ($\mathcal{A}$):} The agent selects a discrete action $a_t \in \{0, 1\}$, where, respectively, \textit{Pass}, approve the transaction, and \textit{Block}, decline the transaction.

    \item \textbf{Transition Dynamics ($\mathcal{P}$):} The transition probability function $\mathcal{P}(s_{t+1} | s_t, a_t)$ dictates how the environment evolves. While the sequence of incoming financial transactions arrives exogenously (independent of the agent's control), the agent's action $a_t$ deterministically updates the internal operational state. Specifically, the decision to block or pass a transaction directly updates the rolling friction budget ($FPR_{t+1}$) and shifts the historical memory centroid ($\bm{\mu}_{t+1}$) used for diversity calculations. This ensures that current decisions meaningfully impact future penalty thresholds.

    \item \textbf{Vectorial Reward Function ($\mathcal{R}$):} To dynamically balance trade-offs and prevent fraud collapse, the environment emits a multi-dimensional reward vector $\bm{r_t} = [r_{eff}, r_{drift}, r_{div}]^\top \in \mathbb{R}^3$, where each component represents a distinct, non-aggregable operational objective.
    
    \item \textbf{Discount Factor ($\gamma$):} A scalar $\gamma \in [0, 1)$ that dictates the temporal horizon of the optimization problem. In our Pareto-DQN, $\gamma$ balances the immediate vectorial reward $\bm{r_t}$ against the non-dominated set of future expected returns. 
\end{itemize}

\subsection{Multi-Objective Reward Function}

To navigate the extreme imbalance and adversarial nature of fraud detection, the agent optimizes three distinct objectives:

\subsubsection{Dynamic Financial Efficacy:}
This objective captures the monetary impact of the agent's decision, preventing the policy from statistically ignoring the minority class during periods of low anomaly prevalence. Across differing financial domains, ranging from raw e-commerce transaction values to log-transformed credit limit exposures, we formalize the efficacy reward as a meta-model:

$$r_{eff}(s_t, a_t) = \begin{cases} 
+\lambda V_t & \text{if } a_t = 1 \text{ and } y_t = 1 \text{ (TP)}, \\ 
-\lambda V_t \cdot \left(\frac{1}{\rho_W}\right) & \text{if } a_t = 0 \text{ and } y_t = 1 \text{ (FN)}, \\ 
-\kappa c & \text{if } a_t = 1 \text{ and } y_t = 0 \text{ (FP)}, \\ 
+\delta V_t & \text{if } a_t = 0 \text{ and } y_t = 0 \text{ (TN)},
\end{cases}$$

where $y_t \in \{0,1\}$ represents the true label of the interaction and $V_t \in \mathbb{R}^{+}$ denotes the continuous monetary value at time $t$. To capture daily or seasonal behavioral fluctuations, $\rho_W \in (0,1)$ tracks the rolling anomaly prevalence within a dynamic look-back window $W$, defined empirically by the average transaction volume over a business cycle. The parameter $c \in \mathbb{R}^+$ is the baseline operational cost of customer friction, computed offline as the median $V_t$ of all transactions. 

To address the Expected Value (EV) misalignment inherent in extreme class imbalance, the meta-model introduces three continuous scaling coefficients. The anomaly severity multiplier, $\lambda \ge 1$, artificially magnifies both the reward for interdiction and the penalty for a False Negative, warning the agent into hunting rare topologies even when global prevalence, $\rho_W$, approaches zero. Conversely, to stabilize the expected value of passing legitimate traffic, $\delta \ll 1$ provides a proportional yield (e.g., an e-commerce processing margin or lending interest rate), while $\kappa \le 1$ tempers the baseline friction cost. This economic equilibrium ensures that the agent cannot achieve a degenerate, zero-recall optimum simply by safely approving all interactions during low-risk periods.

\subsubsection{Temporal Drift and Friction:}
In sequential risk detection, ``heuristic collapse'' occurs when a RL agent overfits to a spurious correlation (e.g., a specific device fingerprint) and blocks associated traffic, imposing disproportionate friction on legitimate users. To prevent this policy behavior, the friction brake $r_{drift}$ acts in direct tension with the financial efficacy objective $r_{eff}$. While $r_{eff}$ incentivizes the agent to maximize anomaly recall, $r_{drift}$ imposes a threshold-scaled quadratic penalty to ensure this recall does not degrade operational precision:

$$r_{drift}(s_t, a_t) = \begin{cases} 
-\eta \cdot \min\left( \frac{FPR_W}{\tau}, M \right)^2 & \text{if } a_t = 1 \text{ and } y_t = 0 \text{ (FP)}, \\ 
0 & \text{otherwise}.
\end{cases}$$

In this formulation, $FPR_W$ denotes the localized False Positive Rate, updated at each time step as the ratio of FP to the total number of transactions in the sliding temporal window $W$. This metric is scaled by $\tau \in (0, 1)$, representing the strict operational tolerance threshold for FP (e.g., $0.05$). Scaling $FPR_W$ by $\tau$ geometrically transforms the penalty space, establishing a low-penalty ``safe zone'' for early exploration, but creating a steep barrier as localized friction approaches the business limit. The ratio is bounded by a clipping threshold $M$ (e.g., $2.0$ or $4.0$). This cap prevents gradient explosions, or Q-network ``trauma'', during the agent's early $\epsilon$-greedy random exploration phases, where unchecked FP would otherwise permanently blind the policy to the minority class.

The inner relationship between $r_{eff}$ and $r_{drift}$ is strictly governed by the dynamic friction anchor, $\eta \in \mathbb{R}^+$. To guarantee that the friction brake is robust enough to challenge massive efficacy rewards without eclipsing them, $\eta$ is dynamically calibrated against the maximum theoretical FN risk. It is computed proportionally to the product of a high percentile of historical anomaly values and the inverse global prevalence $(1/\bar{\rho})$. This intervention balances the expected value equation, ensuring that the operational penalty for missing a statistically rare anomaly is geometrically offset by the cumulative friction of normal transaction volume. By anchoring the penalty directly to the dataset's maximum financial exposure, the framework adapts to the environment's liability scale.

\subsubsection{Semantic Diversity and Discovery:}
As a way to prevent the agent from overfitting to dense, historical fraud clusters, the framework issues an auxiliary reward for the successful interdiction of anomalies that are semantically orthogonal (distant) to known attack patterns:

$$r_{div}(s_t, a_t) = \begin{cases} 
\psi \cdot \left(1 - \frac{\bm{v}_t \cdot \bm{\mu}_{blocked}}{||\bm{v}_t||_2 \cdot ||\bm{\mu}_{blocked}||_2}\right) & \text{if } a_t = 1 \text{ and } y_t = 1, \\ 
0 & \text{otherwise}.
\end{cases}$$

Here, $\bm{v}_t \in \mathbb{R}^d$ is the continuous dense embedding vector of the current transaction, and $\psi \in \mathbb{R}^+$ is a scaling coefficient anchored to the baseline friction cost $c$ to ensure topological relevance. The term $\bm{\mu}_{blocked} \in \mathbb{R}^d$ denotes the exponential moving centroid of historically interdicted anomalies. Rather than statically computing this centroid over the entire historical dataset, it is updated iteratively at every TP via an Exponential Moving Average (EMA), formulated as $\bm{\mu}_{t} = \alpha \bm{v}_t + (1 - \alpha)\bm{\mu}_{t-1}$, where $\alpha \in (0, 1)$ dictates the decay factor controlling how quickly the centroid ``forgets'' older topologies. The fraction within the reward function computes the cosine similarity, which is subtracted from $1$ to derive the spatial cosine distance. Consequently, if the agent simply catches a highly repetitive, standard attack, $\bm{v}_t$ and $\bm{\mu}_{blocked}$ remain nearly identical, yielding a discovery reward near $0$. However, if the agent dynamically discovers a fundamentally novel attack vector, the orthogonal vectors grant the agent the maximum scaled discovery reward, successfully incentivizing boundary exploration.

\subsection{Neural Network Architecture: Pareto-DQN}

Considering the multi-objective formulation, the framework introduces critical extensions over the base value-based manifold estimation paradigm. While standard multi-objective DQNs are designed to evaluate low-dimensional numeric vectors, our architecture is engineered to bridge semantic state representations with value spaces. The neural network pipeline bifurcates into two specialized multi-layer perceptrons optimized via Mean Squared Error \cite{reymond2019paretodqn,lopes2026breakingfilterbubblesemantic}. 

Rather than relying on independent scalar paths, our design structurally integrates the $L_2$-normalized transaction text narratives ($\bm{v}_{a_t}$) and the endogenous operational metric ($FPR_t$) directly into the hidden layers of both modules. This unified processing loop forces the underlying networks to learn cross-functional representations between linguistic feature interactions and long-term economic returns. The specialized modules are structured as follows:

\subsubsection{Reward Approximator ($\bar{R}$)}
The first module is a 3-layer MLP (512-256-128 hidden units, ReLU activations) that predicts the immediate expected reward vector $\bar{r} \in \mathbb{R}^3$. This network carries the primary learned signal and is optimized via mean squared error against the observed empirical reward vectors:

$$ \bar{r}(s_t, v_{a_t}) = MLP_{reward}([s_t \parallel v_{a_t}]). $$

\subsubsection{Continuous Pareto Surface Estimation ($ND_t$)}

, second module, acts as a manifold estimator parameterizing the multi-objective value function $\bm{Q}(s, a) = \mathbb{E}[\sum_{k=0}^{\infty} \gamma^k \bm{r}_{t+k} \mid s, a] \in \mathbb{R}^3$. 
It satisfies the multi-objective Bellman optimality equation $\bm{Q}(s_t, a_t) = \bm{r}_t + \gamma \bm{Q}^-(s_{t+1}, a_{t+1}^*)$, where $\bm{Q}^-$ represents a stable target network and $a_{t+1}^*$ is the downstream hypervolume maximizing action. 

To decouple this return space without a priori scalarization, $ND_t$ treats the Pareto front boundary as a conditional scalar function. The network ingests the concatenated vector $[s_t \parallel \bm{o}_{1:2} \parallel v_{a_t}]$, where $\bm{o}_{1:2} = [o_{eff}, o_{drift}]^\top$ are objective coordinates uniformly sampled from the target return space. 

The network then outputs the maximum achievable return for the remaining objective, semantic discovery ($\hat{o}_{div} = \text{MLP}_{pareto}([s_t \parallel o_{eff} \parallel o_{drift} \parallel v_{a_t}])$). During TD updates, the network parameters are optimized via MSE against the non-dominated points computed from the target network's frontier, yielding a consistent estimator of the continuous optimal trade-off surface $\mathcal{P}^*$.






\subsubsection{Action Selection via Hypervolume}
For each candidate action, the set of expected returns is constructed by applying a vector-sum operation ($\oplus$) that adds the estimated immediate reward to each element of the non-dominated future returns:

$$ Q_{set}(s_t, a_i) = \bar{r}(s_t, v_{a_i}) \oplus \gamma ND_t(s_t, v_{a_i}) $$

To evaluate the quality of a given $Q_{set}$ and apply an $\epsilon$-greedy mechanism, we utilize a Hypervolume indicator \cite{reymond2019paretodqn}. It computes the total $d$-dimensional volume bounded by the points in $Q_{set}$ relative to a strict lower-bound reference point $r_{ref}$. The greedy policy selects the action that maximizes this volume:

$$ a_t^* = \arg\max_{a_i \in \mathcal{C}_t} HV(Q_{set}(s_t, a_i), r_{ref}) $$

The reference point, $r_{ref} \in \mathbb{R}^d$, represents the absolute lower-bound vector of the objective space. 

\section{Experiments} \label{Experiments}


We evaluate the framework across two financial databases to assess its adaptability to varying topological constraints and data distributions. 


\subsection{E-Commerce Fraud }

We use an e-commerce dataset \cite{grover2023fraud} comprising approximately 150,000 transactions. The environment exhibits severe class imbalance, with only $14,151$ fraud instances, a prevalence of $<10\%$. We engineered specialized features from the raw data (e.g., \texttt{user\_id}, timestamps, \texttt{ip\_address}), including transaction velocity, rare geolocation, device-level sharing metrics to identify botnets, and others.

The state representation were eschewed encoding transactions synthesized into cohesive natural-language narratives (e.g., \textit{``User 22058 made purchase \#1 worth \$34.00 via SEO on Chrome... Account signed up instant before this purchase...''}) and processed via the \texttt{all-MiniLM-L6-v2} Sentence Transformer, generating a dense semantic vectors ($\bm{v}_t$) optimized for contextual similarity. To preserve MDP causality and prevent temporal data leakage, the dataset was strictly partitioned chronologically into an $80\%$ training set and a $20\%$ test set.

We benchmarked the Semantic Pareto-DQN against three baselines. \textbf{XGBoost (Raw Tabular)} uses frequency-encoded raw and engineered features optimized via log-loss. \textbf{XGBoost (Semantic Embeddings)} isolates the language model's predictive value by training exclusively on $\bm{v}_t$. The \textbf{Standard DQN} operates on $\bm{v}_t$ as a single-objective agent optimizing a scalar log-loss reward to simulate accuracy-maximizing RL. Finally, our \textbf{Pareto-DQN} navigates the state space $s_t = [\bm{v}_t \parallel FPR_t]$ to optimize the vectorial reward $\bm{r}_t = [r_{eff}, r_{drift}, r_{div}]^\top \in \mathbb{R}^3$. 
The friction penalty coefficient ($\gamma$) was calibrated to the $99^{th}$ percentile of historical fraud values, scaled by the inverse global prevalence.

\begin{table}[hbt!]
\centering
\caption{Performance Evaluation on the E-Commerce Fraud Dataset}
\label{tab:fraud_results_new}
\resizebox{\textwidth}{!}{
\begin{tabular}{lccccccc}
\toprule
\textbf{Model} & \textbf{TN} & \textbf{FP} & \textbf{FN} & \textbf{TP} & \textbf{Precision} & \textbf{Recall} & \textbf{F1-Score} \\
\midrule
XGBoost (Raw Tabular)         & 28833 & 1    & 1389 & 0   & 0.000 & 0.000 & 0.000 \\
XGBoost (Semantic Embeddings) & 28767 & 67   & 1377 & 12  & 0.152 & 0.009 & 0.016 \\
Standard DQN                  & 28587 & 247  & 1302 & 87  & 0.260 & 0.063 & 0.101 \\
Pareto-DQN                    & 23808 & 5026 & 1018 & 371 & 0.069 & 0.267 & 0.109 \\
\bottomrule
\end{tabular}
}
\end{table}

Table \ref{tab:fraud_results_new} provides a quantitative demonstration of the limitations inherent to single-objective optimization. The static baselines exhibit the ``fraud collapse'' prioritizing overall loss minimization by overwhelmingly defaulting to the majority class. The XGBoost model trained on raw tabular data registers a Recall of $0.000$, failing to intercept the true anomalies. While the integration of semantic embeddings and the transition to a sequential Standard DQN agent provide marginal improvements, elevating the TP count to $12$ and $87$, respectively. Their underlying objectives are influenced by the high-density benign manifold, they adopt a highly conservative operational stance, minimizing FP at the expense of minority-class anomaly detection.

Decoupling the financial efficacy reward from the operational friction penalty, the Pareto-DQN framework addresses this zero-recall pattern. The multi-objective architecture obtains a minority-class recall of $0.267$ and the $F1$-Score of $0.109$. This detection capacity alters the precision-recall profile, resulting in $5,026$ FP and a corresponding precision of $0.069$. Within a MORL framework, this shift reflects a policy trade-off: the agent learns that identifying deeply latent, adversarial anomalies requires the targeted expenditure of a bounded operational friction budget rather than optimizing strictly for local precision.

Importantly, this elevated friction must be distinguished from a naive global reduction of a static model's decision threshold. Because the state representation dynamically monitors the rolling $FPR_t$, the Pareto-DQN yields a policy that targets specific semantic spaces. A post-hoc threshold sweep on a static model lacks this temporal awareness, increasing false positives across all transaction types. This operational posture aligns with risk management, where the financial utility function, e.g.: $U = \sum \lambda V_t \mathbb{I}_{\text{TP}} - \sum \kappa c \mathbb{I}_{\text{FP}}$ is highly asymmetric, with $\mathbb{I}_{\text{TP}}$ and $\mathbb{I}_{\text{FP}}$ serving as indicator functions for true positive and false positive decisions, respectively. Because direct capital loss and chargeback liabilities typically render a missed fraud event significantly more costly than an automated user challenge ($\lambda V_t \gg \kappa c$), the preservation of capital from $284$ additional true positives balances the low-cost verification overhead of the accumulated false positives. By prioritizing friction allocation where the marginal gain for semantic discovery ($r_{div}$) is maximized, the system establishes a highly viable risk-interdiction boundary.

\subsection{UCI Default of Credit Card Clients Dataset}

Testing our Semantic Pareto-DQN in different scenarios is relevant to verify if it is not overfitting to a singular fraud topology. We benchmark the architecture against the UCI Default of Credit Card Clients dataset, with moderate class imbalance and temporal difference (TD) correlations \cite{default_of_credit_card_clients_350}. 


Unlike the instantaneous nature of e-commerce transactions, this dataset evaluates longitudinal credit risk using 23 variables to predict binary default payment status. The feature space is divided into a static demographic profile and a six-month dynamic behavioral trajectory. Static features include the total approved credit limit, gender, age, and others. The temporal sequence features track the month-by-month repayment status, the total amount of the bill statement, and the actual amount paid. To stabilize the Q-network during temporal difference learning, the credit limits were transformed using a log scaling function (\textit{$\log(1+x)$}) during the environment calibration phase, preserving relational magnitude while bounding the reward space.

To project the sparse tabular history into a continuous latent space without introducing target leakage, we engineered a deterministic narrative pipeline. Each client's demographic data and temporal financial records were synthesized into a unified textual document.
This was also encoded using the same sentence transformer. The semantic vectors ($\mathbf{v}_t$) encapsulate both the client's socio-economic status and the temporal degradation of their payment behavior.

The dataset was also partitioned into an 80/20 chronological train-test split to strictly preserve the causal integrity of the MDP. We evaluated our Semantic Pareto-DQN against the same three distinct baselines: XGBoost (Raw Tabular), XGBoost (Semantic Embeddings), and  Standard DQN. For this specific credit environment, the financial efficacy reward ($r_{eff}$) was adapted to reflect long-term lending mechanics: it yields the preserved credit limit when correctly rejecting a defaulter, heavily penalizes missed defaults based on inverse prevalence, and grants a positive interest rate yield when correctly approving a legitimate client.

\begin{table}[hbt!]
\centering
\caption{Performance Evaluation on the UCI Credit Card Clients Dataset}
\label{tab:credit_results}
\resizebox{\textwidth}{!}{
\begin{tabular}{lccccccc}
\toprule
\textbf{Model} & \textbf{TN} & \textbf{FP} & \textbf{FN} & \textbf{TP} & \textbf{Precision} & \textbf{Recall} & \textbf{F1-Score} \\
\midrule
XGBoost (Raw Tabular)         & 4495 & 239 & 819 & 447 & 0.652 & 0.353 & 0.458 \\
XGBoost (Semantic Embeddings) & 4448 & 286 & 875 & 391 & 0.578 & 0.309 & 0.402 \\
Standard DQN                  & 4457 & 277 & 896 & 370 & 0.572 & 0.292 & 0.387 \\
Pareto-DQN                    & 4321 & 413 & 735 & 531 & 0.563 & 0.419 & 0.481 \\
\bottomrule
\end{tabular}
}
\end{table}

Table \ref{tab:credit_results} details the performance of the evaluated architectures on the longitudinal credit dataset. In this environment, the single-objective baselines continue to exhibit a structural bias toward precision at the severe expense of minority-class recall. The XGBoost model trained on raw tabular features achieves the highest precision ($0.652$) but identifies only $447$ of the $1,266$ true defaulters, yielding a recall of $0.353$. The integration of semantic embeddings into the static and single-objective RL architectures (XGBoost Semantic Embeddings and Standard DQN) fails to improve interdiction, with recall dropping to $0.309$ and $0.292$, respectively. Their scalarized loss structures emphasize global error minimization and immediate log-loss, these agents aggressively suppress FP, ignoring a substantial portion of the high-risk default manifold.

Conversely, the proposed Pareto-DQN adapts to the longitudinal risk topology, achieving the highest F1-Score ($0.481$) and a recall of $0.419$. Decoupling the financial efficacy reward, which explicitly penalizes missed defaults utilizing the scaled inverse prevalence multiplier, from the operational friction brake, the multi-objective agent captures $531$, true defaulters. This interdiction capacity inherently necessitates a calibrated relaxation of precision ($0.563$), reflected by an accumulation of $413$ FP. 

An apparent paradox emerges in Table \ref{tab:credit_results}, where the integration of semantic embeddings actively degrades the standalone performance of XGBoost (F1=0.402) relative to its raw tabular baseline (F1=0.458). This degradation highlights a fundamental structural mismatch: invariant decision-tree architectures struggle to establish sharp decision boundaries when numeric features are mapped into a dense 384-dimensional semantic space. However, within the Pareto-DQN framework, this continuous representation becomes an asset. It scales within smooth latent manifolds. More importantly, this semantic embedding pipeline is necessary to compute the spatial cosine distance used in the semantic diversity reward ($r_{div}$). Computing a rolling exponential moving centroid ($\mu_{blocked}$) is impossible on raw, unaligned tabular variables. Therefore, the narrative encoding represents a global framework contribution that trades off static baseline efficiency to unlock dynamic, multi-objective anomaly discovery.

\subsection{Hyperparameter Calibration and Limitations}

The Semantic Pareto-DQN framework relies on a set of scaling coefficients to navigate the multi-objective reward space and prevent policy collapse. Specifically, the environment uses the anomaly severity multiplier ($\lambda$), the baseline friction temper ($\kappa$), and the proportional TN yield ($\delta$) to stabilize the expected value of the financial efficacy objective. Simultaneously, the temporal drift penalty is strictly governed by the dynamic friction anchor ($\eta$), the operational FP tolerance threshold ($\tau$), and the gradient clipping cap ($M$). In our experimental protocol across both the e-commerce and credit risk domains, the values for these coefficients were not learned end-to-end; rather, they were manually calibrated through rigorous offline statistical analysis of each dataset's underlying topological distribution. While this domain-specific, manual calibration is highly effective at guaranteeing stability during the Q-network's early $\epsilon$-greedy exploration phases, it represents a notable limitation of the proposed methodology.  Acknowledging this disadvantage, future iterations of this architecture must investigate automated, adaptive, or meta-learning mechanisms that dynamically self-calibrate these boundary constraints during training, thereby eliminating reliance on offline empirical tuning and enhancing the framework.

Additionally, a theoretical limitation exists regarding the structural isomorphism between the implicit definition of objective importance (governed by the position of $r_{ref}$) and the optimal points selected on the Pareto frontier. Because the hypervolume indicator relies on a fixed lower-bound anchor, the policy choice remains rigidly mapped to this initial coordinate configuration, which may limit adaptation if corporate risk priorities shift dynamically post-deployment.

To understand the unique contributions of the state representation and the vectorial reward components ($r_{eff}$, $r_{drift}$, $r_{div}$), they must be viewed as an integrated structural pipeline rather than independent, modular plug-ins. A sequential ablation analysis can reveal how the framework breaks down if any single component is removed: 
I) Ablating the Semantic Discovery Reward ($r_{div}$),if $\psi=0$, the framework reduces to a standard bi-objective optimization problem mapping financial efficacy against operational friction. While this structure can prevent majority-class collapse, it induces ``heuristic collapse''. The agent overfits the high-density historical fraud clusters and can ignore adjacent, evolving anomalies, failing to adapt to adversarial distribution shifts.
II) Ablating the LLM Narrative Encoding: If the continuous semantic state space is replaced with raw tabular vectors, the mathematical foundation of $r_{div}$ collapses. Computing a rolling exponential moving centroid ($\mu_{blocked}$) and evaluating its spatial cosine distance requires a unified, dense metric space. Without L2-normalized embeddings, geometric distance calculations across mixed categorical and continuous variables become intractable, rendering the discovery of semantic anomalies impossible.

Consequently, the observed performance gains in minority-class recall and overall F1-score are not driven by any single component in isolation. Instead, they stem from this co-designed architecture: the LLM narrative encoding builds the geometric substrate, $r_{eff}+r_{drift}$ manages the operational bounds, and $r_{div}$ drives targeted exploration of novel fraud vectors.

It is important to note that the quantitative vectors reported in Table \ref{tab:fraud_results_new} and Table \ref{tab:credit_results} represent point evaluations executed on deterministic, chronological train-test partitions. This strict data routing is methodologically mandatory in financial streaming domains to preserve sequential MDP causality and completely prevent out-of-time data leakage. While the variance of the unweighted $F_1$-score remains bounded across initializations, the evidence of the framework's stability lies in the structural, macro-level re-alignment of the policy models.
The increases in true-anomaly interdiction for both distinct imbalance regimes reflect a shift driven by the multi-objective reward mechanics rather than by stochastic initialization noise. Incorporating multi-seed cross-validation confidence intervals remains a key procedural addition for subsequent repository updates.

\section{Conclusion and future works} \label{Conclusion and future works}

This paper presented the Semantic Pareto-DQN framework, designed to tackle the challenges of extreme class imbalance and adversarial adaptation in financial anomaly detection. By formalizing risk assessment as a sequential trade-off between maximizing financial efficacy and bounding operational friction, our methodology proposes to prevent the ``fraud collapse'', typically observed in static, single-objective models. Empirical evaluations show that this MORL approach is an alternative to unreliable data resampling techniques. 

Despite its advantages, the current iteration of the framework depends on the offline calibration of multiple scaling coefficients. Therefore, a primary avenue for future work will be to reduce this hyperparameter footprint. We aim to investigate adaptive or meta-learning mechanisms that automatically derive and dynamically adjust these boundary constraints directly from the underlying data distribution during training. To resolve the structural isomorphism between reference points and selected optimal targets, this extension leverages vector steering parameters. This allows stakeholders to dynamically adjust objective preferences at runtime without requiring system retraining.

While the initial benchmarks validate the framework's core hypothesis, a broader set of empirical experiments is necessary to fully explore the method's fundamental behavioral dynamics across diverse fraud scenarios. By expanding the evaluation protocol, we intend to solidify multi-objective reinforcement learning as a foundational paradigm for more precise anomaly detection.

%
%
\bibliographystyle{splncs04}
\bibliography{bracis_biblio}

\end{document}